# Analyzing Impact of Socio-Economic Factors on COVID-19 Mortality Prediction Using SHAP Value

Redoan Rahman[1], Jooyeong Kang[2], Justin F Rousseau[3*], Ying Ding[4*]
[1,2,4]School of Information, The University of Texas at Austin, Austin, Texas, United States;
[3]Dell Medical School, The University of Texas at Austin, Austin, Texas, United States

**Abstract**

*This paper applies multiple machine learning (ML) algorithms to a dataset of de-identified COVID-19 patients provided by the COVID-19 Research Database. The dataset consists of 20,878 COVID-positive patients, among which 9,177 patients died in the year 2020. This paper aims to understand and interpret the association of socio-economic characteristics of patients with their mortality instead of maximizing prediction accuracy. According to our analysis, a patient's household's annual and disposable income, age, education, and employment status significantly impacts a machine learning model's prediction. We also observe several individual patient data, which gives us insight into how the feature values impact the prediction for that data point. This paper analyzes the global and local interpretation of machine learning models on socio-economic data of COVID patients.*

**Introduction**

The COVID-19 pandemic has impeded and altered the ways of human life. Its impact has made COVID-19 one of the primary focuses of research since 2019. The disease is responsible for the death of 850,608 people in the USA alone[14]. In the face of a new danger, the knowledge of the deterministic factors of the COVID-19 pandemic has the potential to save lives. For this reason, it is vital to understand the socio-economic characteristics of COVID-19 afflicted patients and determine the correlation between a patient's social and economic condition and their mortality due to COVID-19. The first step to determining the existence of such a correlation is to observe and understand the relationship between a patient's socio-economic characteristics and their mortality due to COVID-19.

This paper examines a dataset containing information regarding patients suffering from the SARS-CoV-2 Coronavirus in 2020. The dataset was constructed by retrieving patients' COVID-19 from a more extensive private database, Covid19ResearchDatabase[19]. The database is a public-private consortium organized by Datavant, Health Care Cost Institute, Medidata, Mirador Analytics, Veradigm, Change Healthcare, SAS, Snowflake, and many others. The filtered dataset contains 20,878 COVID-19 patients, among which 9,177 patients died in 2020. We apply multiple machine learning algorithms to the dataset to understand the impact of the features on the models' prediction. We use the Shapley score to determine each feature's importance in the learning process of Extreme Gradient Boosting (XGBoost) and Random Forest models. Shapley score is a concept recently taken from cooperative game theory and applied to machine learning[18]. We also do individual patient-by-patient prediction analysis, which allows us to observe the impact of specific feature values on the predictions. Finally, we analyze multiple partial dependence plots of features with a higher effect on prediction to understand the trend of its impact on a model's prediction and the feature's interaction with other features.

The findings of this research can become a foundation for determining the governing socio-economic characteristics of COVID. If a feature impacts the prediction model heavily, further research on that feature's impact may discover a correlation between that feature and mortality due to COVID-19. As such, this process can serve as a hypothesis generation task, leading to further studies to establish the causality of specific features on the outcome of interest.

**Related Works**

The impact of COVID-19 has inspired many works of literature in different disciplines. The field of computational learning has also participated in the research to understand and display the impact of COVID-19. In some pieces, the authors have proposed a hybrid machine learning method to predict the progress of the pandemic[1]. The authors utilized data from Hungary to exemplify the model's potential. In another piece of literature, the authors proposed a machine learning-based model using prediction of pandemic impacts and people's subsequent journey and fuel usage of US to project gasoline demand and studied the influence of government intervention[2]. In a different publication, the authors used a convolutional neural network (CNN) to identify the disease and predict outcomes in COVID-19 patients using X-rays and computed tomography (CT)[3]. Some projects have also been where possible applications of machine learning algorithms and methods to battle the COVID-19 pandemic were studied[4]. These applications include analysis of unlabeled data as an input source for COVID-19 and predicting risk in healthcare during the COVID-19 crisis.

\* Joined last authors

Interpretability in Artificial Intelligence (AI) shows promise in healthcare analysis since it allows humans to understand the functionality and process and boosts confidence in the applications. Thus, Shapley value analysis for machine learning model interpretation in healthcare has become more prevalent over the past few years. Using machine learning, researchers proposed a data mining classifier system to predict generalized anxiety disorder (GAD) among women[5]. The authors used Naïve Bayes, Random Forest, and J48 models as classifiers. They demonstrated an apparent increase in model accuracy, sensitivity, and specificity by using Shapley analysis as a feature selection method. Researchers also analyzed a medical imaging dataset to observe the phenomena of different data instance values and showed that an approximation of Shapley value based on a k-nearest neighbors (KNN) classifier could value large amounts of data appraisal within an acceptable amount of time[6]. They also demonstrated other applications of Shapley value such as removing data based on their influence on model learning and detecting noisy labels. A Shapley regression framework has also been proposed as a reliable feature importance evaluator in non-linear models and used the random forest model for demonstration[7].

Previous research sought to predict mortality due to COVID-19. Researchers developed an XGBoost machine learning model to predict the mortality of COVID-infected patients based on biomarkers in blood samples and identified a clinical rule for COVID-19 prognostic prediction[8]. Another piece of literature compares six different models with XGBoost and used Shapley values to feature importance in predicting mortality[9]. The authors apply a "what-if" analysis to determine the impact of marginal changes in the mortality factors on the prediction. However, both studies concentrated on the blood samples of COVID-infected patients and focused on biological features contributing to mortality. This paper focuses primarily on the socio-economic factors influencing mortality in COVID cases.

There have been several studies that attempt to shed light on the disparity in COVID-19 countermeasures. Several authors report that African American people and, to a lesser extent, Latino individuals bear a lopsided affliction of COVID-19-related results and present us with multiple research questions that can provide us with a clear answer to understand the disproportionality[15]. Another group combines the data emphasizing specific risks among marginalized and under-resourced communities, including people in prison or detention centers, immigrants and the undocumented, people with disabilities, and homeless people in USA[16]. Researchers also assess disparity in COVID-19 vaccine distribution in multiple countries across the world[17]. They fit a logistic model to report daily case numbers and estimated the vaccine roll-out index (VRI). Then they used a Random Forest model and analyzed the relation between predictors and model prediction. The authors found that median per capita income, human development index, the proportion of internet users in the population, and health expenditure per capita are the top four factors associated with VRI. These studies concentrate on analyzing and determining the socio-economic factors that impact patient care. In our work, we focus more on analyzing socio-economic factors and their association with mortality due to COVID-19.

**Methodology**

A.  Extreme Gradient Boosting (XGBoost)

XGBoost is part of a family of Boosting algorithms. It is an end-to-end Tree boosting algorithm that is sparsity-aware, can be used for sparse data and estimated tree learning, and is scalable10. It offers various tuning parameters that make it suitable for different scenarios.

The scalability of XGBoost is primarily due to the algorithmic optimization in multiple stages of the tree learning process. XGboost uses an approximate algorithm that suggests possible splitting points based on feature distribution percentiles. The algorithm aggregates the statistics of the continuous features split by the candidates and finds the best solution based on the statistics.

B.  Random Forests

Random forest is a machine learning technique that utilizes ensemble learning. The algorithm consists of many tree predictors. The outcome of a random forest model depends on each prediction of all predictor trees[11]. Generally, the average or mean of the trees' output is the algorithm's final prediction.

C.  SHAP (SHapley Additive exPlanations) Values

SHAP is an integrated framework that provides some tools to interpret predictions. It assigns a value to each feature in a computational learning model that represents its importance. [12] proposed SHAP values as a unified measure of feature importance. It connects LIME, DeepLIFT, and later-wise relevance propagation method with Shapley regression and Shapley sampling.

*1) Shapley Value*

---

\* Joined last authors

The feature importance for linear models in the presence of multicollinearity is known as the Shapley regression value or Shapley value[13]. It signifies the effect of including that feature on the model prediction. If the feature impacts the model positively, then the assigned Shapley value to the feature is positive, and if the effect is negative, then the Shapley value reflects that impact.

*2) SHAP Feature Importance Plot*

The SHAP feature importance plot illustrates the relative importance of the features where large absolute Shapley values are globally important. The algorithm uses the average of absolute Shapley values per feature to demonstrate the level of impact of the features in model prediction. The calculated absolute feature importance is plotted in descending order to create the SHAP feature importance plot.

*3) SHAP Summary plot*

Each point in the SHAP summary plot is a Shapley value for a feature. The feature determines the vertical position of the point, and the Shapley value determines the horizontal position. The color of the point represents whether the value of the feature is low or high. Our experiment uses red and blue to represent low or high feature values, respectively. For example, for a feature Age, an older man would be drawn as red or a redder point, whereas a younger would be described as blue or a bluer point. Overlapping points are jittered in the y-axis position. The SHAP summary plot indicates a possible relationship between feature value and the impact on model prediction. However, it does not prove any causal relationship.

*4) SHAP Partial Dependence Plot*

In SHAP partial dependence plot, individual feature values are plotted on the X-axis, and corresponding Shapley values are plotted on the Y-axis. This plot displays:

- The relationship between the feature and the target value demonstrates whether the relation is linear, straightforward, monotonic, or more complex.
- The relation with another feature based on the frequency of inter-feature interaction.

D. Dataset Statistics & Feature Description

We use a filtered dataset retrieved from the Covid19ResearchData database for our experiments. The database had three different schemas titled 'ANALYTICSIQ', 'OFFICE_ALLY', and 'MORTALITY'. OFFICE_ALLY provides claims and remittance data from a claims clearinghouse. MORTALITY or Death Index is a curated set of records from multiple obituary data sources. ANALYTICSIQ contains data at the name/geographic level. It is aggregated from large public data sources including census, econometric data from the US government, summarized credit data from 2 different credit bureaus, home sales information from county courthouses, occupation information from state licensing boards, and past purchase behavior from catalogers and retailers et cetera. Features from ANALYTICSIQ have inferred characteristics based on consumer and custom survey data.

The data from these schemas were joined based on common identifiers. The patients in the database had one or more diagnosis codes from claims from OFFICE_ALLY. To determine if a patient has tested positive for COVID-19, we looked through all diagnosis codes and selected all patients with a diagnosis code containing the string 'U07.1', 'U07.2', 'J12.82' or 'M35.81'. We then used the patient's birth date and the appointment date to calculate their age. We also removed any patient data if they did not have the necessary feature information. 20,878 patients had a diagnosis of COVID-19 in the dataset, among which 9,177 patients died in 2020. The others are presumed as alive for the experiments. For the experiments, we used the following features:

- INCOMEIQ_PLUS_V3: Indicates a household's predicted annual income (ANALYTICSIQ).
- INCOMEIQ_PLUS_DISP_V2: Represents a household's predicted disposable income (ANALYTICSIQ).
- ETHNICIQ_V2: This element identifies an individual's ethnicity, known and inferred (ANALYTICSIQ).
- AIQ_EMPLOYMENT: This element predicts an individual's employment status on a scale from 1-7, where 1-3 means a very low likelihood of employment, 4 means part-time job, and 5-7 means full-time employment (ANALYTICSIQ).
- AIQ_EDUCATION_V2: Denotes an individual's level of education from less than high school through a graduate degree (ANALYTICSIQ).
- HW_ER_VISITS_SC: Prediction of likelihood of having visited the emergency room (ER) in the last 12 months (7=most likely; 1=least likely) (ANALYTICSIQ).

- HW_PRIMARY_CARE_DOCTOR_SC: Prediction of likelihood of having a primary care doctor (7=most likely; 1=least likely) (ANALYTICSIQ).
- HW_MED_UTILIZATION: This element predicts if an individual is likely to exhibit high medical utilization by visiting three or more of the following in the last 12 months: emergency room, medical specialist, primary care doctor, urgent care. (Y/N) (ANALYTICSIQ).
- HW_STRESS_V2: Represents the predicted measure of stress levels (7-high stress; 1=low stress) (ANALYTICSIQ).
- AGE: Indicates an individual's age (OFFICE_ALLY).
- MORTALITY: This element is the target variable for the learning model. It represents whether the patient died due to COVID-19 or not (0=non-mortality or survived 1=mortality or death) (MORTALITY).

E. Experiment Setup & Task Description

Records were only included if data were available in the MORTALITY dataset. This resulted in the final curated dataset having a 44% mortality rate. However, our focus for the experiment was to train well-performing machine learning models and identify the most impactful socio-economic factors for the trained models using explainable AI. While observing the prediction of deleterious and neutral phenotypes of human missense mutations in human genome data, [20] observed that balanced training data (50% neutral and 50% deleterious) results in the highest balanced accuracy (the average of True Positive Rate and True Negative Rate. For this reason, we attempted to achieve a more balanced dataset rather than replicating real-world scenarios. This approach also helped us avoid duplicating or oversampling any data class.

We used Snowflake for accessing the Covid-19ResearchDatabase and used python for data preprocessing, model training, and Shapley analysis. We trained a Random Forest and an XGboost model using the abovementioned features. We split the data into 90% and 10% for training and testing, respectively. Also, since we are using categorical data, we used the OrdinalEncoder scaling technique provided by the python sklearn library to prepare the data for model training.

For the XGBoost model, we used the XGBRegressor provided by the xgb python library. We used 'reg:logistic' as the objective since we are working on a classification problem. We also used 0.1 for learning_rate, 5 for max_depth, and 10 for n_estimators.

For the Random Forest model, we used the RandomForestRegressor provided by the python sklearn library. We use 6 for max-depth and 10 for n_estimators while setting up the model for training.

After the models had been trained, we used the shap library to create the feature importance plot and summary plot to understand the impact of the features on model prediction. We also made several partial dependence plots to understand a feature's effect on model prediction more clearly and to understand the feature's interaction with other features. Finally, we conducted individual case-by-case analyses to understand the local importance of features.

**Result & Observations**

In this section, we present the results of the experiments and the observations we made from the results.

A. Feature Importance

Our experiments produced SHAP feature importance plots and summary plots for both XGBoost and Random Forest models. These plots demonstrate the following information:

- Feature importance: The features are ranked in descending order. Therefore, the feature plotted at the top in the SHAP feature importance plot and the SHAP summary plot are the most impactful on the model's prediction.

- Impact: The horizontal location of the individual plotted point in the SHAP summary plot demonstrates feature importance or impact on model prediction.

- Feature Value: Each plotted point represents an instance or a data point. The point's color shows whether the feature value is in a higher (redder) or lower (bluer) value for that instance.

- Relationship: The preceding information can reveal a possible relationship between the target result, whether the patient died in 2020 or not, and the feature as detailed below. However, the inferred relationship does not act as evidence of real-world causality.

* Joined last authors

1) Feature Importance observation for the XGBoost model

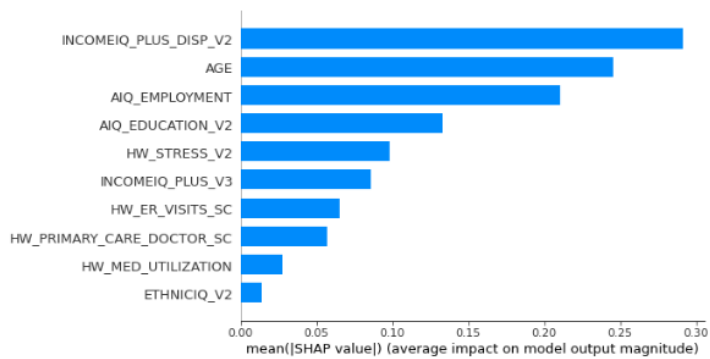

**Figure 1.** SHAP Feature Importance Plot of XGBoost model

Figure 1 displays the average impact of each feature on model output. According to the figure, the disposable income of households has the most impact on the production of the XGBoost model. The patient's age, employment, education status, and stress level also play a decisive role in the output. The likelihood of ER visits, the possibility of exhibiting heavy med utilization, the chance of having a primary care doctor, or the patient's ethnicity has little effect on the output prediction of XGBoost.

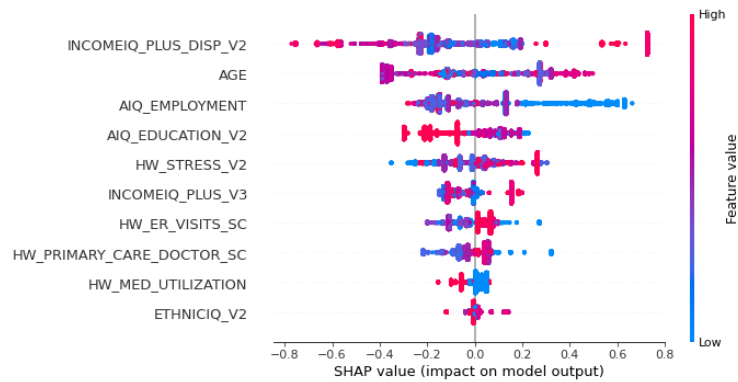

**Figure 2.** SHAP Summary Plot of XGBosst model

Figure 2 displays the SHAP value for each feature instance and its impact on the model. From the color of the points and their horizontal positions, we can reach several conclusions.

Figure 1 shows that disposable income level, age, and employment are the top 3 highly ranked features based on the mean SHAP value. But Fig 1 B indicates that only the employment feature has the apparent decisive power that COVID patients with lower employment have a higher probability of dying. At the same time, disposable income level and age are unclear because COVID patients with high disposable income and age are distributed across all ranges of the mean SHAP scores, including having a higher probability of dying or not dying. Education ranked No four, and its SHAP score distribution shows that COVID with low education has a high certainty of dying because all the blue dots are located at the positive SHAP scores. Even Medical Utilization ranked No nine but has an apparent decisive power that COVID patients with high medical utilization have a higher chance of not dying. In summary, income, age, employment, education, stress, ER revisit, primary care, medical utilization, and ethnicity contribute to the mortality prediction for COVID patients. Among these critical features, income, age, and employment are the top 3 factors that significantly contribute to the final mortality. Employment, education, and medical utilization have clear decisive patterns.

2) Feature Importance observation for the Random Forest model

Fig. 3 displays the feature importance plot for the random forest model. According to the figure, unlike the XGBoost model, the annual income of the patient's households has the most impact on the output of the random forest model. The patient's employment, education status, age, and household's disposable income also play a decisive role in the prediction. Stress level, the likelihood of ER visits, the possibility of exhibiting heavy med utilization, the likelihood of having a primary care doctor, or the patient's ethnicity has little effect on the output prediction of random forest.

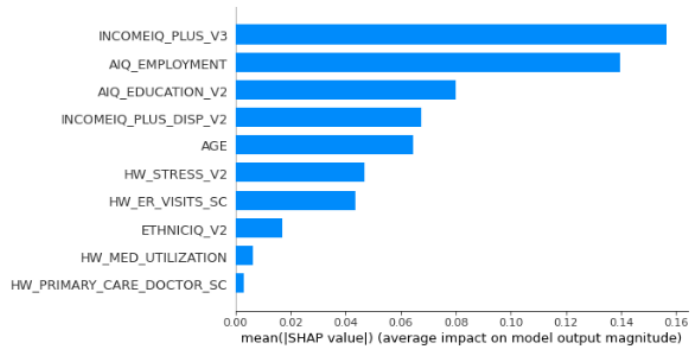

**Figure 3.** SHAP Feature Importance Plot of Random Forest model

Fig. 4 displays the SHAP value for each feature value and their impact on the output of the random forest model. We can reach several conclusions from the color of the points and their horizontal positions.

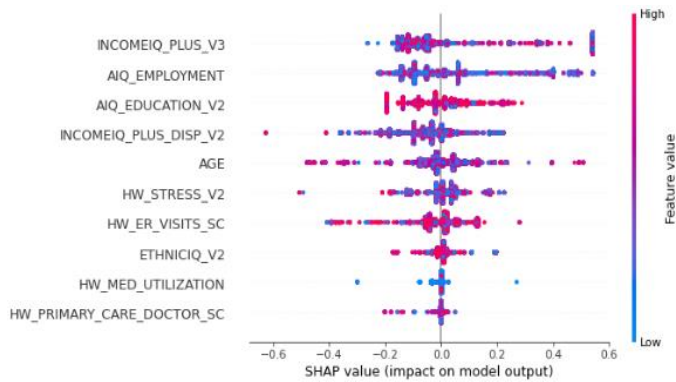

**Figure 4.** SHAP Summary Plot of Random Forest model

According to the Random Forest Model, Figure 3 shows that income level, employment, and education are the top 3 highly ranked features based on the mean SHAP value. But Fig 4 indicates that only the education feature has the roughly apparent decisive power that COVID patients with lower education have a higher probability of dying because COVID patients with low education are concentrated on the positive SHAP scores indicating a higher probability of dying. Different from Figure 1, age and disposable income level is ranked No 5 and 4. But looking at Fig 4, almost all features have unclear decision patterns because each feature's high and low values are distributed across all ranges of the SHAP scores, including having a higher probability of dying or not dying.

We can observe that the SHAP summary plot of the XGBoost model provided a clearer understanding of the feature impact on model prediction than the random forest summary plot. However, whether this is related to inherent characteristics of the random forest model or SHAP value analysis or whether this is related to the experiment's setup cannot yet be confirmed and requires further analysis.

By comparing the features from the XGBoost and Random Forest models (see Fig 1(A) and Fig 2(A)), we found that disposable income level, age, education, employment, stress, medical utilization, and ethnicity contribute significantly to the mortality prediction of COVID patients. From Fig 1b and Fig 2b, we can conclude that employment and education have contributed significantly to the mortality prediction of COVID patients. Both have the clear decisive patterns that COVID patients with low education and low employment have higher risks of dying.

* Joined last authors

B. Partial Dependence Plot of Features

We also produced several partial dependence plots of features to understand feature impact on model output.

1) Partial dependence plots for the XGBoost model

a) Education

Figure 5 displays the partial dependence plot between the patient's disposable household income and its assigned SHAP values. The relationship is a complex one. We can see that push to increase the prediction outcome. The increase in income decreases the SHAP value, indicating that higher income decreases the prediction outcome.

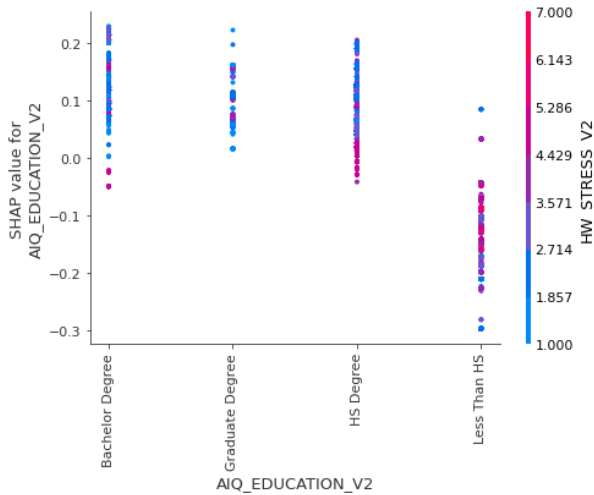

**Figure 5.** Partial Dependence plot for patient's education level for XGBoost model

b) Employment

Figure 6 shows the partial dependence plot between patients' employment level and their morality prediction. As the employment value increases, we can see that the SHAP value becomes more negative, implying that better employment status leads to a decreased prediction of mortality value. Employment feature most frequently interacts with disposable income, and the income is lowest when the employment level is lowest.

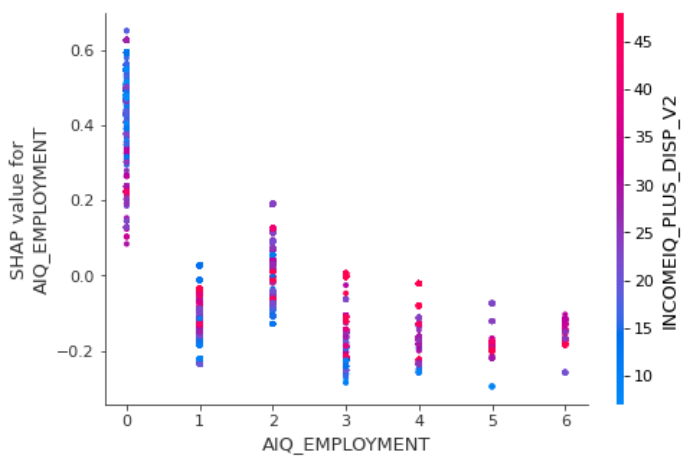

**Figure 6.** Partial Dependence plot for patient's employment level for XGBoost model

* Joined last authors

2) Partial dependence plots for the Random Forest model

a) Age

Figure 7 displays the partial dependence plot for the patient's age. We can see that as age increases, the likelihood of having a primary care doctor increases. As for the impact on prediction, the figure does not display any clear relation between age value and SHAP value. We have observed a similar result in Fig. 4. However, the instances with an age value higher than 60 have a higher SHAP value. This phenomenon implies that for patients with higher age, age impacts the model prediction positively, which means higher age has more chance of outputting a death prediction.

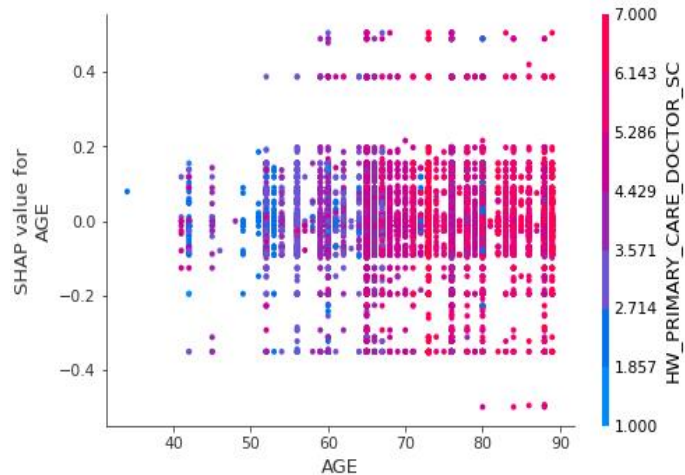

**Figure 7.** Partial Dependence plot for patient's age for Random Forest model

b) Disposable Household Income:

Figure 8 shows the partial dependence plot for the patient's disposable household income. From the figure, we can say that when an increase in employment level occurs, a parallel increase in disposable income also occurs. However, similar to the impact of annual income, the patient's disposable household income does not have a clear impact on the model prediction.

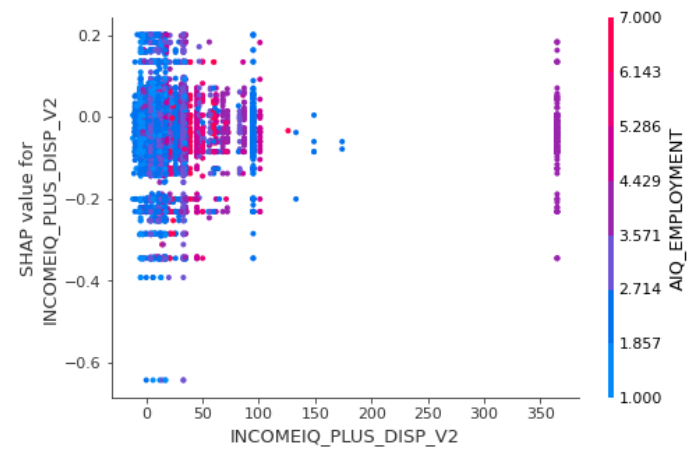

**Figure 8.** Partial Dependence plot for patient's Disposable Household Income for Random Forest model

C. Local Interpretation

This section looks at several individual cases and how the feature values impacted the output through multiple individual SHAP value plots.

In Figure 9, we see an individual SHAP value plot. This individual case was processed through the XGBoost model. The lower education value, higher stress value and ethnicity indicate the higher mortality risk, whereas the higher annual and disposable income of the household and higher employment value pushes contribute to the higher survival rate. We can see that ethnicity plays a significant role in this prediction, even though the SHAP summary plot and feature importance plot ranked ethnicity as a minor impactful feature, but in this individual case, ethnicity is the top ranked feature contributing the most to mortality rate.

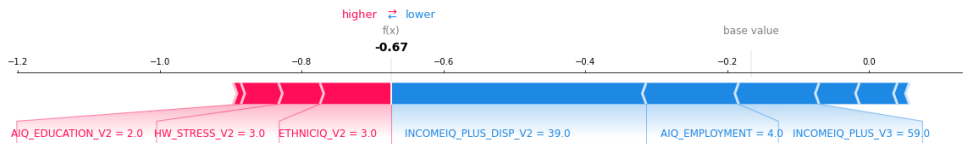

**Figure 9.** Partial Dependence plot for patient's education for XGBoost model

In Figure 10, there is another individual SHAP value plot, but this case was processed through the random forest model. The higher stress value contributes significantly to the mortality risk. The higher annual and disposable income value and the higher education value are influencing the mortality prediction to be lower.

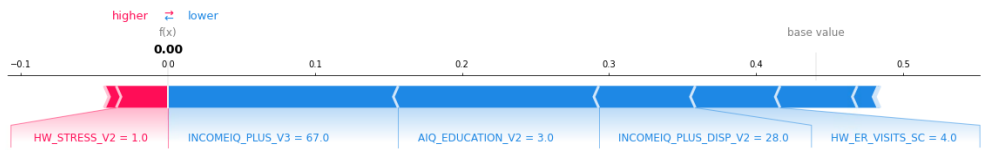

**Figure 10.** Partial Dependence plot for patient's education for Random Forest model

**Limitation & Future Work**

This research is preliminary work toward understanding the impact of socio-economic factors in healthcare. The experiment conditions are controlled resulting in several limitations. These data should be interpreted in the context of the study design. Given a dataset and applying an ML prediction model to predict mortality, we sought to understand the socioeconomic drivers of the predictive models. The performance of the prediction model could only be applied in the context of the constraints of the dataset. Caution should be made to generalize our findings to the population. But the methods we employed could be applied to prediction models on more representative datasets.

Our work considers a dataset that has a 44% mortality rate. This indicates these data are not representative of the population and contain bias in their selections. This limits the generalizability of our findings, and we seek in future work to apply these methods to more representative data. We did not include non-Covid patient data for our experiment, further making the generalizability of our work unclear. We intend to extend our dataset to include the non-Covid patient as well so that we can identify the similarities and differences of the impactful characteristics between Covid and non-Covid patients.

We are currently aiming to introduce different explainable AI methods in our experiments and analyze the results. Our purpose is to determine if the results stay unchanged. If the results vary, then we also are aiming to understand the reasons behind the differences.

**Conclusion**

This paper analyses the possible socio-economic factors that may impact mortality due to covid. There were 20,878 COVID patients in the dataset, among which 9,177 patients died. We trained an XGBoost and a Random Forest model and later used SHAP value analysis to understand the impact of the features on the output of the models.

In our research, we focused on interpretability and worked on the assumption that if a feature impacts a model's learning heavily, then it is an excellent indicator to decide if a patient is at high risk for mortality due to COVID. We found that a patient's household disposable and annual income, employment, education level, and patient's age are good possible indicators to determine if a patient should be considered high-risk or not. We also found that a patient's stress level, er visit frequency, and doctor visit frequency are also adequate indicators. We also observed these behaviors and trends through partial dependence plots and individual SHAP value plots. However, some characteristics show impactful behavior in the local interpretation, such as ethnicity, but are ranked low in the global interpretation. We will require further analysis to understand the implications of these findings in the context of other studies[21] showing the impact of these features while also considering the limitations of the data.

These findings should be verified through further data analysis and case studies before these can be used in decision-making in the real world. Further research on these findings may help us identify the population in danger of mortality due to COVID. It has the potential to assist in clinical decision support to understand the impact of features on risk prediction at the individual level and the population level thus also supporting public health policy. It may also help develop a quick triage method to determine priority for COVID-infected people.

* Joined last authors